\documentclass[biblatex,english]{template/lni}
\usepackage{caption}
\usepackage{listings}
\usepackage{subcaption}
\usepackage{booktabs}
\usepackage{breakurl}
\usepackage{multirow}
\usepackage{pdflscape}
\usepackage{afterpage}
\usepackage{capt-of}% or use the larger `caption` package
\usepackage{rotating}

\DeclareFontFamily{OT1}{mathc}{}
\DeclareFontShape{OT1}{mathc}{m}{n}{ <-> mathc10 }{}

\newbibmacro{string+url}[1]{%
\iffieldundef{url}{% no url -- see whether there is a doi
  \iffieldundef{doi}{#1}{\href{https://dx.doi.org/\thefield{doi}}{#1}}%
  }{\href{\thefield{url}}{#1}}}
\DeclareFieldFormat*{title}{\usebibmacro{string+url}{#1}}

\AtEveryBibitem{
    \ifboolexpr{
        not test {\ifentrytype{online}} and
        not test {\ifentrytype{software}}
      }
        {\clearfield{url}}
    {}
    
}

\usepackage[]{blindtext}

\begin{document}
%%% Mehrere Autoren werden durch \and voneinander getrennt.
%%% Die Fußnote enthält die Adresse sowie eine E-Mail-Adresse.
%%% Das optionale Argument (sofern angegeben) wird für die Kopfzeile verwendet.
\title[Distillation Techniques for Document Understanding]{Leveraging Distillation Techniques for Document Understanding: A Case Study with FLAN-T5}
%%%\subtitle{Untertitel / Subtitle} % falls benötigt
\author[1]{Marcel Lamott}{marcel.lamott@hs-rm.de}{0009-0009-4345-6888}
\author[2]{Muhammad Armaghan Shakir}{raoarmaghanshakir040@gmail.com}{0009-0004-3476-2772}
%\author[1]{Vorname3 Nachname3}{vorname3.name3@affiliation1.de}{0000-0000-0000-0000}
%\author[1]{Vorname4 Nachname4}{vorname4.name4@affiliation1.de}{0000-0000-0000-0000}%
\affil[1]{RheinMain University of Applied Sciences\\LAVIS\\Unter den Eichen 5\\65195 Wiesbaden\\Germany}
\affil[2]{National University of Sciences and Technology (NUST)\\Scholars Ave, H-12, Islamabad, Islamabad Capital Territory\\Pakistan}
\maketitle

\begin{abstract}
The surge of digital documents in various formats, including less standardized documents such as business reports and environmental assessments, underscores the growing importance of Document Understanding.
While Large Language Models (LLMs) have showcased prowess across diverse natural language processing tasks, their direct application to Document Understanding remains a challenge.
Previous research has demonstrated the utility of LLMs in this domain, yet their significant computational demands make them challenging to deploy effectively. Additionally, proprietary Blackbox LLMs often outperform their open-source counterparts, posing a barrier to widespread accessibility. 
In this paper, we delve into the realm of document understanding, leveraging distillation methods to harness the power of large LLMs while accommodating computational limitations. Specifically, we present a novel approach wherein we distill document understanding knowledge from the proprietary LLM ChatGPT into FLAN-T5. Our methodology integrates labeling and curriculum-learning mechanisms to facilitate efficient knowledge transfer. %, empowering FLAN-T5 to exhibit proficient document comprehension capabilities. 
%Through rigorous experimentation and evaluation, we demonstrate the effectiveness of our approach, highlighting the enhanced performance of FLAN-T5 in document understanding tasks.
This work contributes to the advancement of document understanding methodologies by offering a scalable solution that bridges the gap between resource-intensive LLMs and practical applications. Our findings underscore the potential of distillation techniques in facilitating the deployment of sophisticated language models in real-world scenarios, thereby fostering advancements in natural language processing and document comprehension domains.
\end{abstract}


\section{Introduction}
% X Warum ist Document Understanding wichtig
% X Warum LLMs für Document Understanding nutzen
% X Warum sollte man Distillation nutzen
% Latin prompt erwähnen
Natural language processing (NLP) has rapidly advanced, with language models becoming increasingly capable of understanding and generating human language~\cite{zhao2023survey}.
But this development has come at a cost: recent language models, such as Falcon~\cite{almazrouei2023falcon} and Llama~\cite{touvron2023llama}, have become increasingly large, putting their training out of reach for many institutions and individuals.
As such, the distillation of knowledge~\cite{hinton2015distilling} from larger, sophisticated models into more compact, efficient ones is a crucial area of research.
This process, known as knowledge distillation, aims to retain the performance of large models while reducing computational costs, making these models more accessible and practical for various applications.
%%%%
Document understanding, a key aspect of NLP, involves comprehending and interpreting rich layouts paired with complex texts~\cite{subramani2021survey}.
Effective document understanding is essential for numerous applications, such as: information retrieval, 
%compliance and risk management, automated customer service, 
legal document processing, 
%financial document analysis and 
medical record analysis, etc.
%Enhancing models for document understanding can lead to more intuitive and efficient tools, benefiting both businesses and individuals.
Enhancing models for document understanding thus has an impact on a wide variety of domains. % and can lead to more intuitive and efficient tools, benefiting both businesses and individuals.
%%%%
Curriculum learning, proposed by Bengio et al.~\cite{curriculumlearning}, is inspired by the human educational process and has emerged as a powerful strategy to enhance the training of machine learning models. 
In curriculum learning, the training data is presented to the model in a meaningful order, typically progressing from simpler to more complex examples. 
This approach can improve convergence rates and overall performance by aligning the learning process with the natural progression of cognitive skill acquisition. 
By systematically evaluating different curriculum learning strategies, this research aims to optimize the distillation process from a large language model into a more streamlined one.

In this work we focus on the LLM-centric document understanding pipeline presented in~\cite{lamott2024lapdoc}, which combines the text with document layout:
From a document's OCR data, a purely textual document representation is generated, which aims to preserve the original layout information.
These document representations are then combined with task specific prompt templates and fed into an LLM, which solves the document understanding task at hand.
This approach offers the benefit of simplicity, as it requires no model fine-tuning. Moreover, it utilizes the reasoning capabilities and knowledge capacity of LLMs.
This study focuses on transferring knowledge from a state-of-the-art language model, namely ChatGPT 3.5~\cite{liu23chatgpt}, into a more efficient model of the FLAN-T5 series~\cite{FLANt5}, using supervised fine-tuning, labeling for knowledge elicitation and leveraging various curriculum learning approaches.
FLAN-T5 is chosen as the student model because of its high performance in relation to its small size.

The goal of this study is to enhance document understanding capabilities while maintaining performance and reducing resource demands.
%By exploring and comparing these strategies, we aim to contribute to the development of more effective and accessible NLP technologies. 
Overall, we make the following contributions:

\begin{enumerate}
    \item A set of comprehensive experiments carried out on five different research datasets, comparing the performance of our distillation approach to state-of-the-art models.
    \item A comparison of student models of three different sizes and various learning curricula, providing insight on viable training strategies.
\end{enumerate}

% X Was machen wir
% Forschungsfragen
% X Contributions

\section{Related Work}
% Document UNderstanding allgemein
% DOcument Understanding mit LLMs
% LLMs allgemein
% Distillation allgemein
% Task specific distillation

% LLMS
\textbf{LLMs}:
%LLMs utilizing the attention-based transformer architecture~\cite{transformer} represent a significant advancement in NLP.
%These models typically comprise encoder-decoder architectures, where the encoder processes input data and the decoder generates output sequences. 
%While early LLMs favored encoder-only architectures, recent trends show the rise of decoder-based models as the dominant approach. LLMs with decoder architectures, exemplified by the GPT (Generative Pre-trained Transformer) series, demonstrate impressive capabilities in generating coherent and contextually relevant text across various tasks and demonstrate notable reasoning capabilities, especially in a zero-shot setting~\cite{kojima2023large}. 
%To facilitate utilizing these abilities, instruction tuning~\cite{zhang2023instruction} is used as an additional training step:
%While a decoder-LLM's goal is to predict the next word of a sequence, instruction tuning trains the LLM to follow human instructions.
%Instruction tuning has also been shown to improve model performance and generalization to unseen tasks, as is the case for the student model used in this work: FLAN-T5~\cite{FLANt5}, the instruction-tuned version of T5~\cite{t5}.
LLMs utilizing the attention-based transformer architecture~\cite{transformer} represent a significant advancement in NLP, as they demonstrate impressive capabilities in generating coherent and contextually relevant text across various tasks and demonstrate notable reasoning capabilities, especially in a zero-shot setting~\cite{kojima2023large}.
Instruction tuning~\cite{zhang2023instruction}, i.e. training the LLM to follow human instructions, has been shown to improve model performance and generalization to unseen tasks, as is the case for the student model used in this work: FLAN-T5~\cite{FLANt5}, the instruction-tuned version of T5~\cite{t5}.

%Accordingly, we focus on instruction-tuned decoder models in this work.

% Document understanding with multi modal models:
% Muss die erwähnen weil ich sie referenziere in evaluation:
% LayoutLMv2, LayoutLMv3, UDOP
\textbf{Multimodal Models}: 
Multimodal models in document understanding integrate text, images, and layout information. The LayoutLM series~\cite{layoutlm, layoutlmv2, layoutlmv3} utilizes a BERT-type transformer encoder~\cite{bert}, which processes a concatenation of word embeddings and visual patch embeddings. UDOP~\cite{udop} follows a generative approach, reconstructing text layout through an encoder-decoder model. These models achieve notable results on various benchmarks, demonstrating the impact of multimodal integration on complex document understanding tasks.

\textbf{LLMs for Document Understanding}:
Recent works have aimed to exploit the superior knowledge capacity and reasoning capabilities of LLMs for document understanding.
Fujitake et al. and Luo et al. \cite{fujitake2024layoutllm, luo2024layoutllm}
extend an existing LLM architecture with a vision encoder from a pretrained document understanding model, such as LayoutLM~\cite{layoutlm}.
LAPDoc and LATIN-Prompt~\cite{lamott2024lapdoc, latinprompt} forego model architecture changes and fine-tuning and instead utilize the generalist problem solving capabilities of LLMs via incorporation of layout information into task specific prompts.
%  encoder that encodes document images and a decoder that interprets tasks, and outputs.
% Mit finetuning: 
% https://arxiv.org/pdf/2403.14252
% https://arxiv.org/pdf/2404.05225
% Ohne finetuning: (LLMs can serve as generalist agents for ad-hoc problem solving, without fine-tuning to specific tasks.)
% LATIN
% LAPODOC

% Distillation
% Curriculum learning
% Layout Prompts: LAPDoc & LATIN Prompt

\textbf{Knowledge Distillation}:
Knowledge distillation, introduced by Hinton et al.~\cite{hinton2015distilling}, compresses large, complex models (teacher models) into smaller, efficient ones (student models). This process has been widely applied across various domains, particularly in NLP, to reduce computational overhead while maintaining high levels of performance. 
Recent studies like~\cite{sanh2020distilbert} and~\cite{jiao-etal-2020-tinybert} have shown the effectiveness of distilling large pre-trained transformers into smaller versions, demonstrating competitive performance with reduced resource requirements.
Xu et al.~\cite{xu2024survey} explore the distillation of LLMs in particular and highlight the wide variety of applications and approaches.
%They list different methods of knowledge elicitation, i.e. sources for the distillation training data, among them \emph{labeling}, where the teacher model labels the student model's training data.

\textbf{Curriculum Learning}:
Proposed by Bengio et al.~\cite{curriculumlearning}, curriculum learning draws inspiration from human learning processes, suggesting that models can learn more effectively when trained on data that is organized from simple to complex. This approach has been shown to improve the convergence speed and generalization ability of machine learning models and has been applied to various tasks in the context of NLP, such as question answering~\cite{xu-etal-2020-curriculum} and machine translation~\cite{platanios-etal-2019-competence}.
%In the context of NLP, curriculum learning has been applied to various tasks, including language modeling and machine translation. For example, Platanios et al. (2019) introduced competence-based curriculum learning for neural machine translation, demonstrating improved translation quality and training efficiency.

\section{Approach}
% LAPDoc SpatialLayout formatted Prompts
% in ChatGPT 3.5 füttern, output ist label
% lora_FLAN_t5 small base large
%As the aim of the distillation setting is to transfer knowledge from the teacher into the student, we train the student models not with the original labels for the training data.
%Instead, the teacher model labels the data for the student:
%To this end, we convert each document into a textual representation using LAPDoc SpatialLayout verbalization strategy~\cite{lamott2024lapdoc}.
To label the training data for the student, we convert each document into a textual representation using LAPDoc SpatialLayout verbalization strategy~\cite{lamott2024lapdoc}.
These textual representations are then inserted into task specific prompt templates and fed into the teacher LLM to generate the targets for our student model.
During training of the student model, we utilize curriculum learning and present the training data in order of increasing difficulty to the student with the goal to aid its convergence and generalizability.

%We format the documents with LAPDoc \texttt{SpatialLayout} verbalization strategy, which is similar to LATIN-Prompt.

%The approach is divided into two parts: 
%First, knowledge elicitation where data for the distillation of the teacher into the student is generated. 
%Second, curriculum learning where the student is trained on the data generated by the teacher, while the data is presented to the student in order of increasing difficulty.

%\red{Es ist nirgendswo klar das JSON ausgegeben wird}

\subsection{Knowledge Elicitation}\label{subsec:knowledge}
\begin{figure}
\centering
\includegraphics[width=\textwidth]{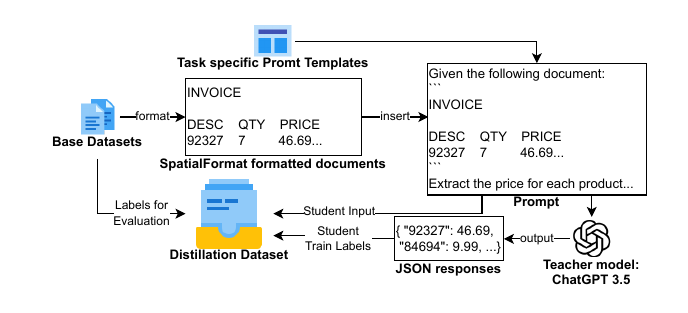}
\caption{Overview of the knowledge elicitation: the documents from the base datasets are converted to a textual representation with LAPDoc SpatialFormat verbalizer and inserted into task specific prompt templates. The prompts are used as input for the distillation dataset and to generate the training labels by feeding them into the teacher LLM: ChatGPT 3.5.\protect\footnotemark}
\label{fig:knowledge}
\end{figure}
%\red{Wieso verwenden wir keine Labels?}

\textbf{Labeling}:
\footnotetext{Icons from \url{https://icons8.com}}
Figure \ref{fig:knowledge} shows an overview of our labeling approach to generate the distillation training data:
Given a document and its corresponding OCR data, we retrieve a purely textual document representation by using LAPDoc SpatialFormat verbalization.
This verbalization strategy aims to reconstruct the documents original layout by formatting the bounding box contents via insertion of spaces and newlines.
The textual document representation is then inserted into a prompt template together with task specific instructions, e.g. key information to extract from the document.
The prepared prompt is then fed into the teacher LLM and its output serves as target for the student.
Each prompt asks for the results to be delivered as JSON.\footnote{\url{https://www.json.org/}}

\textbf{Prompt Templates}:
We use the LAPDoc prompt templates without format description~\cite{lamott2024lapdoc}.
While T5 models usually receive task specific prefixes in their prompts, we forego these prefixes and instead fine-tune the model for the new task.

\subsection{Curriculum Learning}\label{subsec:learning}
\begin{figure}
\centering
\includegraphics[width=\textwidth]{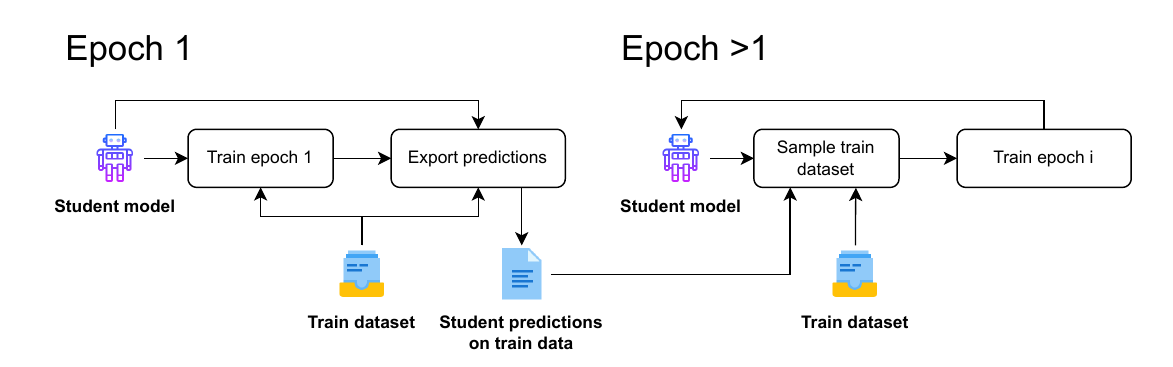}
\caption{Overview of the curriculum learning: the student is trained for one epoch, after which its predictions are exported and used to generated the datasets for subsequent epochs. These sampled datasets present the data in order of increasing difficulty based on the previous predictions of the student.\protect\footnotemark}
\label{fig:curriculum_learning}
\end{figure}
%TODO: motivation temperature sampling erwähnen

% Was wird gemacht
Figure \ref{fig:curriculum_learning} shows an overview of the curriculum learning approach, which is motivated by the temperature sampling applied by language models to diversify their outputs~\cite{chang2023kldivergence}:
Initially, the student model $s$ is trained for one epoch on the train dataset $T$, after which its predictions on the training data
$\Pi = \{\, s(x) \mid x \in T \,\}$ are collected.
In each subsequent epoch $i$, the training data $T_i$ is sampled from $T$ according to sampling probability 
$$p(x) \propto max(0.01, \sigma(\Pi_x,y))^{\tau_i}$$
where $\sigma$ is a similarity measure, $y$ is the label for sample $x$ and $\tau_i \in \mathbb{R}$ is a temperature parameter that is decreased with each epoch.
For $\sigma$ we use normalized Levenshtein distance~\cite{normlevdist} after normalizing both $x$ and $y$, i.e. by removing whitespace in the JSON structure.  
%To avoid discontinuity for negative $\tau_i$, we take $0.01$ as minimum similarity. 
We take $0.01$ as minimum similarity as the domain of $a^b = \{ (a,b) \in \mathbb{R}^2 \mid a \neq 0 \text{ or } b \geq 0\}$.
%Domain of p(a,b)={(a,b)∈R 2 ∣a  =0 or b≥0}
%%
% Was ist die Motivation / der Effekt
This definition of $T_i$ results in the training data being presented in order of increasing difficulty to the student model across the epochs.
\footnotetext{Icons from \url{https://icons8.com}}

%This is motivated by the fact that the task of interpreting SpatialLayout formatted prompts is new to the student.
%Further, some training samples provide very challenging cases due to the complexity of the source document, for which the SpatialLayout formatting approach is of limited utility for the LLM.   
%Thus introducing the new content representation format with simple examples to the student is promising in ensuring its convergence and generalizability. 
% Keine format description

\section{Experiments}
% Brauchen richtige baseline!

In the following experiments, we investigate whether the student model is able to learn layout reasoning and document understanding from the teacher's signal, which is generated using the LAPDoc SpatialFormat approach.
The student's document understanding capabilities are evaluated using a set of research datasets, which evaluate different tasks such as key information extraction and document question answering.
Further, we analyze the impact of different learning curricula on the student model's performance and provide comparisons to state-of-the-art models.

%To investigate layout awareness in depth, we also take a qualitative look at a subset of manually annotated challenge cases (see Section~\ref{subsubsec:sroiequalanalysis}).
%In the following experiments, we investigate whether suitable verbalization strategies can support LLMs with better layout reasoning and provide exemplary comparisons of open-source and commercial solutions.
%In most experiments, we measure the awareness of the LLM towards layout aspects 
%indirectly, i.e. 
%via  accuracy on document understanding tasks (which include research benchmarks and industry datasets, see Section~\ref{subsec:datasets}).
%To investigate layout awareness in depth, we also take a qualitative look at a subset of manually annotated challenge cases (see Section~\ref{subsubsec:sroiequalanalysis}).

\subsection{Datasets and Setup}
%The distillation dataset is generated by feeding the SpatialFormat formatted documents of the base datasets into the teacher LLM and collecting its response, as detailed in Section~\ref{subsec:knowledge}.

%During training of the student model, the distillation dataset is ss

% Was sind denn überhaupt die Datensätze: documents, tasks...

%, from which the distillation dataset is generated as detailed in section~\ref{subsec:knowledge},
\textbf{Base Datasets}: The base datasets include a variety of document understanding tasks:
Visual Question Answering (DocVQA~\cite{docvqa}, InfographicsVQA~\cite{infovqa}), Table Question Answering (WikiTableQuestions~\cite{wtq}), Table Natural Language Inference (TabFact~\cite{tabfact}), Key Information Extraction (SROIE~\cite{sroie}) and Structural Reading Comprehension (WebSRC~\cite{websrc}).
%DocVQA contains scans of various documents and invoices.
%InfographicsVQA consists infographics such as explanatory posters.
%WikiTableQuestions and TabFact are collections of images of Wikipedia tables.
DocVQA, InfographicsVQA, TabFact and WikiTableQuestions are part of the DUE benchmark~\cite{due}, from which we did not include DeepForm~\cite{deepform}, Kleister Charity~\cite{kleister} and PWC~\cite{pwc}, as these datasets contain documents with a very high number of pages and are out of the scope of this work.
%SROIE is a collection of invoice scans, and WebSRC is a set of question-answer pairs collected from various web pages.
WebSRC gives one question per sample, thus we group up to 10 questions referring to the same web page into a single prompt.
In total, the datasets amount to $57202$ train samples and $4329$ test samples.
Table \ref{tab:datasets} in the Appendix shows an overview of the base datasets.

% Avg test split tokens
%DocVQA 518
%InfographicsVQA 756
%TabFact 718
%WikiTableQuestions 1366
%SROIE 399
%WebSRC 483

\textbf{Distillation Dataset}:
%The distillation dataset is generated by the procedure detailed in section~\ref{subsec:knowledge}: For each document in the base datasets, the SpatialFormat formatted prompt is fed into the teacher LLM and also serves as input to the student model. 
%The response from the teacher is saved and serves as target for the student.
For the distillation dataset (see Section~\ref{subsec:knowledge}), we create a validation split $\text{Eval}_{\text{Base}}$ with size of 1\% of the training split.
After the first epoch, the initial distillation dataset is sampled based on each sample's difficulty for the student model (see Section~\ref{subsec:learning}).
During these phases, we create another validation split $\text{Eval}_{\text{This}}$ with size of 1\% of the remaining training split.
Thus, the distillation dataset for the first epoch has $51481$ train, $5721$ eval and $4329$ test samples while the distillation datasets for the remaining epochs have $45760$ train, 5721 $\text{eval}_{\text{base}}$, $5721$ $\text{eval}_{\text{this}}$ and $4329$ test samples.

%\subsection{Setup}

\textbf{OCR}:
Each dataset in the DUE benchmark comes with a selection of pre-applied OCR engines, where we use \texttt{microsoft\_cv} for DocVQA, InfographicsVQA and WikiTableQuestions and \texttt{tesseract} for TabFact. 
Microsoft Computer Vision OCR is used for WebSRC.
SROIE includes OCR data.

\textbf{LLMs}:
% Teacher
For the teacher model we use gpt-3.5-turbo-110613 in JSON mode, with a temperature of 0, and enter each prompt in the role of user. 
% Student
%For the student we evaluate three different sizes of FLAN-T5: small\footnote{\url{https://huggingface.co/google/flan-t5-small}} with 77M paramters, base\footnote{\url{https://huggingface.co/google/flan-t5-base}} with 248M parameters and large\footnote{\url{https://huggingface.co/google/flan-t5-large}} with 783M parameters.
For the student we evaluate three different sizes of FLAN-T5: \textsc{SMALL} with 77M paramters, \textsc{BASE} with 248M parameters and \textsc{LARGE} with 783M parameters.\footnote{\url{https://huggingface.co/google/flan-t5-small}, \url{https://huggingface.co/google/flan-t5-base}, \url{https://huggingface.co/google/flan-t5-large}}
Evaluation of the larger variants \textsc{XL} and \textsc{XXL} has been omitted due to time constraints.
% top_p=0.9
For prediction, student models use parameter $\text{top\_p}=0.9$ and all other parameters with the HuggingFace default values.

\textbf{Student Training}:
All student models are trained with LoRA~\cite{hu2021lora} for a total of $8$ epochs on an NVIDIA A100 SXM4 40GB.
LoRA reduces the number of trained parameters for small to $0.9\%$, for base to $0.7\%$ and for large to $0.6\%$.
All student models use a learning rate of $1\mathrm{e}{-4}$, decreasing to 0 with a linear schedule.
For $\text{FLAN-T5}_{\textsc{LARGE}}$ we use a batch size of 4 and a batch size of 8 for the other models.
%We further use one gradient accumulation step, doubling the effective batch size.

\textbf{Curriculum Learning}:
After the first training epoch, the temperature $\tau_i$ (see section~\ref{subsec:learning}) for epoch $i$ is given by $\tau_i = \mathbb{T}_{\text{start}} + \mathbb{T}_{\text{step}} \cdot (i-1)$, where $\mathbb{T}_{\text{start}}$ and $\mathbb{T}_{\text{step}}$ are hyperparameters governing the course of the temperature during training.
We explore four different choices of temperature parameters $(\mathbb{T}_{\text{start}}, \mathbb{T}_{\text{step}}, \tau_8)$, where $\tau_8$ denotes the temperature in the last epoch: 
$\mathcal{A} = (0.25, -\frac{1}{12}, -0.25)$, 
$\mathcal{B} = (0.5, -\frac{1}{6}, -0.5)$,
$\mathcal{C} = (1, -\frac{1}{3}, -1)$ and
$\mathcal{D} = (2, -\frac{2}{3}, -2)$.
Further, we train a baseline model for each student $\mathcal{O} = (0,0,0)$, which uses the same training dataset for all epochs. 

\textbf{Metrics}:
The evaluation is performed following the procedure presented in Lamott et al.~\cite{lamott2024lapdoc}: For the DUE datasets, we use the official evaluation repository with its given metrics: ANLS for DocVQA and InfographicsVQA and accuracy for TabFact and WikiTableQuestions.
For SROIE, we use a type-aware accuracy measure.
We do not evaluate our approach on WebSRC.

\subsection{Results}\label{subsec:results}
% Verlgeichen Student mit Teacher
% Wie gut hat JSON und Keys hinbekommen?
%The course of the curriculum learning is dependent on the temperature parameters introduced in section~\ref{subsec:learning}.
Apart from the following experiments, we further evaluated each T5 model in its non-fine-tuned version, resulting in a score of $0.0$ across all models and datasets.\footnote{Inspection of the predictions shows, that all models failed 
to follow the task instructions and to generate the requested JSON output format.}

\textbf{Overall Performance}:
Table~\ref{tab:sota} compares the performance of our teacher and the student models to various other works.
It shows that the students achieve promising results on SROIE and TabFact, but overall cannot compete with multimodal models.
%However, FLAN-T5-large shows promising results, indicating that the next larger model in this series, FLAN-T5-xxl, could compete with ChatGPT 3.5.

% \usepackage{multirow}
\begin{table}[]
\begin{tabular}{llclcclcc}
\hline
\multicolumn{2}{l}{\multirow{2}{*}{Model}}                                         & KIE           &                      & \multicolumn{2}{c}{Question Answering} &                      & \multicolumn{2}{c}{Table QA/NLi} \\ \cline{3-3} \cline{5-6} \cline{8-9} 
\multicolumn{2}{l}{}                                                               & SROIE         &                      & DocVQA             & InfoVQA           &                      & WTQ             & TabFact        \\ \hline
\multicolumn{2}{l}{$\text{T5}_{\textsc{LARGE}}+\text{2D}+\text{U}$~\cite{due}}     & -             &                      & 81.0               & 46.1              &                      & 43.3            & 78.6           \\
\multicolumn{2}{l}{$^\dagger\text{LayoutLMv2}_{\textsc{LARGE}}$~\cite{layoutlmv2}} & \textbf{97.8} &                      & \textbf{85.3}      & -                 &                      & -               & -              \\
\multicolumn{2}{l}{$^\dagger\text{LayoutLMv3}_{\textsc{LARGE}}$~\cite{layoutlmv3}} & -             &                      & 83.4               & 45.1              &                      & 45.7            & 78.1           \\
\multicolumn{2}{l}{$^\dagger\text{UDOP}$~\cite{udop}}                              & -             & \multicolumn{1}{c}{} & 84.7               & 47.4              & \multicolumn{1}{c}{} & 47.2            & \textbf{78.9}  \\
\multicolumn{2}{l}{$\text{LATIN-Prompt (Claude)}$~\cite{latinprompt}}              & -             &                      & 82.6               & 54.5              &                      & -               & -              \\ \hline
\multicolumn{2}{l}{{\scriptsize (Teacher) $\text{LAPDoc (ChatGPT 3.5)}$~\cite{lamott2024lapdoc}}}                             & 77.0          &                      & 79.8               & \textbf{54.9}     &                      & \textbf{47.7}   & 70.1           \\
Ours                   & $\text{FLAN-T5}_{\textsc{SMALL},\mathcal{O}}$             & 47.6          &                      & 28.3               & 9.3               &                      & 4.2             & 65.0           \\
Ours                   & $\text{FLAN-T5}_{\textsc{BASE},\mathcal{O}}$              & 66.6          &                      & 49.8               & 14.6              &                      & 8.4             & 68.5           \\
Ours                   & $\text{FLAN-T5}_{\textsc{LARGE},\mathcal{O}}$             & 73.0          &                      & 63.3               & 21.6              &                      & 15.6            & 70.9           \\ \hline
\end{tabular}
\caption{Comparison of our approach to state-of-the-art models. 
It shows that the students achieve promising results on SROIE and TabFact, but overall cannot compete with multimodal models.
\textdagger~denotes that the model uses vision, text and layout modalities, while all other models use only text and layout modalities.
}
\label{tab:sota}
\end{table}

% Vergleichen Temperature training gegen baseline
\textbf{Curriculum Learning}:
The results of the distillation trainings are presented in Table~\ref{tab:students}.
We also compare our students against a baseline denoted by $\mathcal{GT}$, which is trained on the original labels instead of the teachers labels.
It is shown, that document understanding knowledge was successfully distilled into the student models.
While all students had difficulties with InfoVQA and WTQ, $\text{FLAN-T5}_{\textsc{BASE}}$ and $\text{FLAN-T5}_{\textsc{LARGE}}$ managed to surpass the teacher on TabFact, with the latter also achieving close performance on SROIE.
On DocVQA, InfoVQA and WTQ some students even managed to outperform the GT baseline.
Further it is shown, that the larger models benefitted more from the curriculum learning.

% \usepackage{multirow}
% \usepackage{multirow}
\begin{table}[h]
\begin{tabular}{lclcclcclc}
\hline
\multirow{2}{*}{Model}                         & KIE                    &                      & \multicolumn{2}{c}{Question Answering} &                      & \multicolumn{2}{c}{Table QA/NLi}       &                      & \multirow{2}{*}{Avg}     \\ \cline{2-2} \cline{4-5} \cline{7-8}
                                               & SROIE                  &                      & DocVQA             & InfoVQA           &                      & WTQ           & TabFact                &                      &                          \\ \hline
ChatGPT 3.5 (teacher)                          & 77.0                   &                      & \textbf{79.8}      & \textbf{54.9}     &                      & \textbf{47.7} & 70.1                   &                      & \textbf{65.9}            \\ \hline
$\text{FLAN-T5}_{\textsc{SMALL},\mathcal{GT}}$ & \textit{57.49}         & \multicolumn{1}{c}{} & 28.7               & 10.1              & \multicolumn{1}{c}{} & 4.4           & \textit{92.9}          & \multicolumn{1}{c}{} & \textit{38.7}            \\
$\text{FLAN-T5}_{\textsc{SMALL},\mathcal{O}}$  & 47.6                   &                      & 28.3               & 9.3               &                      & 4.2           & 65.0                   &                      & 30.9                     \\
$\text{FLAN-T5}_{\textsc{SMALL},\mathcal{A}}$  & 40.4                   &                      & \textit{30.0}      & 9.4               &                      & \textit{4.5}  & 65.4                   &                      & 29.9                     \\
$\text{FLAN-T5}_{\textsc{SMALL},\mathcal{B}}$  & 35.8                   &                      & 26.2               & \textit{10.2}     &                      & 3.8           & 66.1                   &                      & 28.4                     \\
$\text{FLAN-T5}_{\textsc{SMALL},\mathcal{C}}$  & 43.5                   &                      & 27.1               & 10.0              &                      & 4.2           & 65.0                   &                      & 30.0                     \\
$\text{FLAN-T5}_{\textsc{SMALL},\mathcal{D}}$  & 21.8                   &                      & 25.2               & 9.2               &                      & 3.7           & 67.3                   &                      & 25.4                     \\ \hline
$\text{FLAN-T5}_{\textsc{BASE},\mathcal{GT}}$  & \textit{76.8}          & \multicolumn{1}{c}{} & \textit{51.4}      & 15.5              & \multicolumn{1}{c}{} & 8.9           & \textit{94.5}          & \multicolumn{1}{c}{} & \textit{49.4}            \\
$\text{FLAN-T5}_{\textsc{BASE},\mathcal{O}}$   & 66.6                   &                      & 49.8               & 14.6              &                      & 8.4           & 68.5                   &                      & 41.6                     \\
$\text{FLAN-T5}_{\textsc{BASE},\mathcal{A}}$   & 64.6                   &                      & 50.1               & 15.2              &                      & 9.4           & 68.4                   &                      & 41.5                     \\
$\text{FLAN-T5}_{\textsc{BASE},\mathcal{B}}$   & 63.3                   &                      & 49.9               & \textit{16.6}     &                      & \textit{10.5} & 69.5                   &                      & 42.0                     \\
$\text{FLAN-T5}_{\textsc{BASE},\mathcal{C}}$   & 65.6                   &                      & 47.9               & 15.4              &                      & 8.7           & 65.6                   &                      & 40.6                     \\
$\text{FLAN-T5}_{\textsc{BASE},\mathcal{D}}$   & 56.8                   &                      & 51.3               & 15.6              &                      & 9.0           & 71.3                   &                      & 40.8                     \\ \hline
$\text{FLAN-T5}_{\textsc{LARGE},\mathcal{GT}}$ & \textit{\textbf{86.5}} & \multicolumn{1}{c}{} & \textit{65.1}      & \textit{23.6}     & \multicolumn{1}{c}{} & 15.1          & \textit{\textbf{94.6}} & \multicolumn{1}{c}{} & \textit{57.0}            \\
$\text{FLAN-T5}_{\textsc{LARGE},\mathcal{O}}$  & 73.0                   &                      & 63.3               & 21.6              &                      & 15.6          & 70.9                   &                      & 48.9                     \\
$\text{FLAN-T5}_{\textsc{LARGE},\mathcal{A}}$  & 74.1                   &                      & 63.0               & 20.9              &                      & 15.0          & 70.9                   &                      & 48.8                     \\
$\text{FLAN-T5}_{\textsc{LARGE},\mathcal{B}}$  & 74.1                   &                      & 63.9               & 20.8              &                      & 16.1          & 71.3                   &                      & 49.2                     \\
$\text{FLAN-T5}_{\textsc{LARGE},\mathcal{C}}$  & 71.3                   &                      & 62.1               & 20.2              &                      & 15.0          & 70.1                   &                      & \multicolumn{1}{l}{47.7} \\
$\text{FLAN-T5}_{\textsc{LARGE},\mathcal{D}}$  & 73.7                   &                      & 64.7               & 22.3              &                      & \textit{16.3} & 72.6                   &                      & 49.9                     \\ \hline
\end{tabular}
\caption{Results of knowledge distillation from the teacher, LAPDoc with SpatialFormat (ChatGPT 3.5)~\cite{lamott2024lapdoc}, to our student models. 
$\mathcal{GT}$ refers to the baseline trained with original labels instead of the labels generated by the teacher model.
$\mathcal{O}$ refers to training without curriculum, $\mathcal{A}$, $\mathcal{B}$, $\mathcal{C}$ and $\mathcal{D}$ refer to the curricula listed in Section~\ref{subsec:results}.
Best results across all models are shown in \textbf{bold}, best results for each model are shown in \textit{italics}.
It shows that all students had difficulties with InfoVQA and WTQ, while $\text{FLAN-T5}_{\textsc{BASE}}$ and $\text{FLAN-T5}_{\textsc{LARGE}}$ managed to surpass the teacher on TabFact, with the latter also achieving close performance on SROIE.
%Further it is shown, that the larger models benefitted more from the curriculum learning.
% InfoVQA und WTQ schwer
}
\label{tab:students}
\end{table}

\section{Conclusion}
% Keine format prefixes
% small viel zu klein
% Größere modelle versuchen, wegen time constraints nicht möglich gewesen
% Vielleicht LR scheduling ein problem und deswegen bringt curriculum nicht viel
We distilled document understanding knowledge from ChatGPT 3.5 into FLAN-T5 of three different sizes 
\textsc{small}, \textsc{base} and \textsc{large}
%$\text{FLAN-T5}_{\textsc{SMALL}}$, $\text{FLAN-T5}_{\textsc{BASE}}$ and $\text{FLAN-T5}_{\textsc{LARGE}}$
while employing curriculum learning.
The results are promising:
While $\text{FLAN-T5}_{\textsc{SMALL}}$ struggles with most datasets, its performance on TabFact is surprisingly close to that of the teacher.
$\text{FLAN-T5}_{\textsc{LARGE}}$ even managed to surpass the teacher's performance on TabFact.
%Counterintuitively, some students managed to outperform the GT baseline on DocVQA, InfoVQA and WTQ.
Further, the results suggest that the larger T5 models utilize more pronounced learning curricula better than their smaller counterparts.
%Further, it can be observed that all curricula decreased performance for $\text{FLAN-T5}_{\textsc{SMALL}}$ compared to the baseline $\mathcal{O}$, curriculum $\mathcal{B}$ lead to small improvements for $\text{FLAN-T5}_{\textsc{BASE}}$ and curriculum $\mathcal{D}$ proved beneficial for $\text{FLAN-T5}_{\textsc{LARGE}}$, suggesting that the larger models utilize more pronounced learning curricula better than their smaller counterparts.
A starting point for future work is the evaluation of larger student models, such as $\text{FLAN-T5}_{\textsc{XL}}$ and $\text{FLAN-T5}_{\textsc{XXL}}$ and different learning curricula.
Another point of interest is investigation into other document representations and their utility for knowledge distillation.
%Another point of interest is the evaluation of more sophisticated learning curricula as well as the utilization of different verbalization strategies. %, as the evaluation results suggest, that these could prove competitive to the teacher and may constitute more adequate student models.

%\blindtext
\newpage

%% \bibliography{lni-paper-example-de.tex} ist hier nicht erlaubt: biblatex erwartet dies bei der Preambel
%% Starten Sie "biber paper", um eine Biliographie zu erzeugen.
\printbibliography

\appendix

\section*{Appendix: Datasets}\label{sec:datasets}
The following table shows an overview of the used base datasets:
\begin{table}
\centering
\begin{tabular}{lrrr}
\toprule
Dataset & \#Train & \#Test & Avg. Tokens per Document \\
\midrule
DocVQA & 10194 & 1287 & 518 \\
InfographicsVQA & 4406 & 579 & 756 \\
WikITableQuestions & 1350 & 421 & 1366 \\
TabFact & 13182 & 1695 & 718 \\
SROIE & 626 & 347 & 399 \\
WebSRC & 27444 & 0 & 483 \\
\midrule
 $\Sigma$ & 57202 & 4329 &  \\ %$\varnothing $ 707
\bottomrule
\end{tabular}
\caption{Number of samples for each base dataset. 
%For WebSRC, we group multiple questions into a single prompt, so the reported number of training samples is smaller than the actual number of questions. We also do not use the testsplit for WebSRC.
For WebSRC, we group multiple questions into a single prompt and do not use the testsplit.
%Average tokens for each dataset are counted on the test split with the tiktoken BPE tokenizer, which is provided by OpenAI for use with their models.
Average tokens are counted on test split with OpenAI's tiktoken BPE tokenizer.\protect\footnotemark
}
\label{tab:datasets}
\end{table}

\footnotetext{\url{https://github.com/openai/tiktoken}}

% TODO: tensorboard graphen

\end{document}